\documentclass[10pt,twocolumn,letterpaper]{article}

\usepackage[pagenumbers]{iccv} 

\usepackage{xspace}
\usepackage{amssymb}
\usepackage{amsmath}
\usepackage{amsfonts}
\usepackage{wrapfig}
\usepackage[ruled,vlined]{algorithm2e}
\SetKwInput{KwInput}{Input}
\SetKwInput{KwOutput}{Output}
\usepackage{multicol}
\usepackage{multirow}
\usepackage{makecell}
\usepackage{tabularx}
\usepackage{adjustbox}
\usepackage{pifont}
\usepackage{color}
\usepackage{colortbl}

\definecolor{lightgray}{rgb}{0.8, 0.8, 0.8}
\definecolor{lgray}{rgb}{0.66, 0.66, 0.66}

\definecolor{lblu_tab}{RGB}{225, 235, 246}
\definecolor{orange_vitad}{RGB}{222, 131, 68}
\definecolor{blue_vitad}{RGB}{106, 153, 208}
\definecolor{trajectory_green}{RGB}{126, 171, 85}
\definecolor{trajectory_yellow}{RGB}{245, 194, 66}
\definecolor{tab_others}{RGB}{235, 235, 235}
\definecolor{tab_ours}{RGB}{225, 235, 246}

\definecolor{whit_tab}{RGB}{255, 255, 255}
\definecolor{gray_tab}{RGB}{246, 246, 246}
\definecolor{oran_tab}{RGB}{252, 242, 237}
\definecolor{blue_tab}{RGB}{227, 240, 251}

\newcommand{\cmark}{\ding{52}\xspace}%
\newcommand{\xmarkg}{\textcolor{lightgray}{\ding{56}}\xspace}%
\definecolor{iccvblue}{rgb}{0.21,0.49,0.74}
\definecolor{linkcolor}{RGB}{255,0,0}
\definecolor{urlcolor}{RGB}{255,105,180}
\definecolor{citecolor}{RGB}{66,168,235}
\usepackage[pagebackref,breaklinks,colorlinks,allcolors=iccvblue,linkcolor=linkcolor,urlcolor=urlcolor,citecolor=citecolor]{hyperref}

\def\paperID{4} 
\def\confName{ICCV}
\def\confYear{2025}

\title{Bridge Feature Matching and Cross-Modal Alignment with Mutual-filtering \\ for Zero-shot Anomaly Detection}

\author{Yuhu Bai\textsuperscript{1}\thanks{Equal contribution.}
\quad Jiangning Zhang\textsuperscript{1,2}\footnotemark[1]
\quad Yunkang Cao\textsuperscript{3} 
\quad Guangyuan Lu\textsuperscript{1} \\
\quad Qingdong He\textsuperscript{2} 
\quad Xiangtai Li\textsuperscript{4}  
\quad Guanzhong Tian\textsuperscript{1}\thanks{Corresponding author.} \\
\normalsize \textsuperscript{1}{Zhejiang University} \quad \textsuperscript{2}{YouTu Lab, Tencent} \quad \textsuperscript{3}{Huazhong University of Science and Technology} \quad \textsuperscript{4}{Peking University} \\
{\tt\small Code: \url{https://github.com/ybai111/FiSeCLIP}}
}

\begin{document}
\maketitle
\begin{abstract}
With the advent of vision-language models (e.g., CLIP) in zero- and few-shot settings, CLIP has been widely applied to zero-shot anomaly detection (ZSAD) in recent research, where the rare classes are essential and expected in many applications.
This study introduces \textbf{FiSeCLIP} for ZSAD with training-free \textbf{CLIP}, combining the feature matching with the cross-modal alignment. 
Testing with the entire dataset is impractical, while batch-based testing better aligns with real industrial needs, and images within a batch can serve as mutual reference points. 
Accordingly, FiSeCLIP utilizes other images in the same batch as reference information for the current image. However, the lack of labels for these references can introduce ambiguity, we apply text information to \textbf{fi}lter out noisy features. 
In addition, we further explore CLIP's inherent potential to restore its local \textbf{se}mantic correlation, adapting it for fine-grained anomaly detection tasks to enable a more accurate filtering process. 
Our approach exhibits superior performance for both anomaly classification and segmentation on anomaly detection benchmarks, building a stronger baseline for the direction, e.g., on MVTec-AD, FiSeCLIP outperforms the SOTA AdaCLIP by +4.6\%$\uparrow$/+5.7\%$\uparrow$ in segmentation metrics AU-ROC/$F_1$-max.
%
\end{abstract}    
\section{Introduction} \label{sec:introduction}

Anomaly detection (AD) has found extensive applications in various domains~\cite{ad-survey, cao2024survey, ader, liu2024deep}, where it plays a crucial role. 
AD aims to identify if a sample has any anomalies and to pinpoint the anomalous locations. 
Previous anomaly detection~\cite{rd, patchcore, CAVGA,uniad,pyramidflow} approaches mainly follow an unsupervised paradigm, yet they still rely on a substantial amount of normal samples for training. 
However, requiring substantial computational resources, these methods also lack strong generalization abilities. 
Zero-shot anomaly detection (ZSAD) has emerged~\cite{jeong2023winclip, bzsad}, allowing direct inference without any target domain data, and has recently garnered significant attention from researchers.

\begin{figure}
    \centering
    \includegraphics[width=1\linewidth]{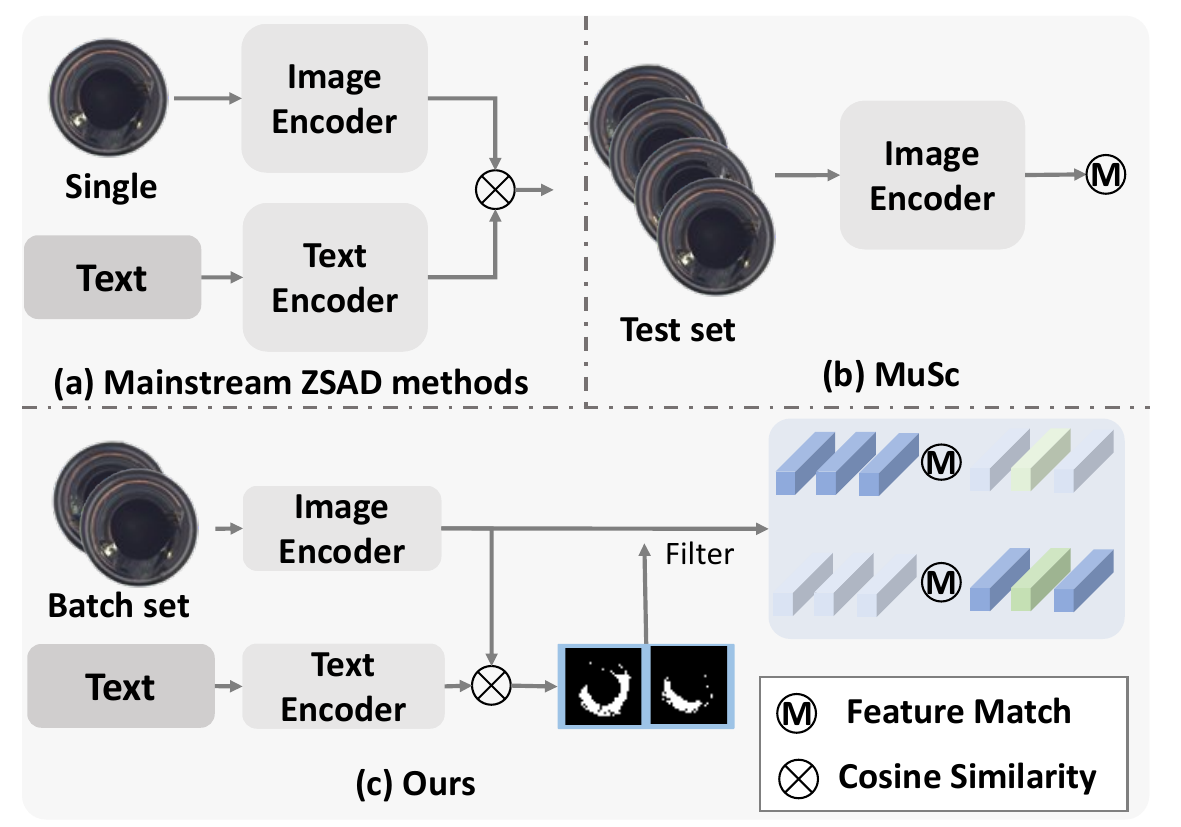}
    \caption{Compared with (a) mainstream ZSAD methods~\cite{jeong2023winclip, april-gan} and (b) MuSc~\cite{musc} method. Our method uses a batch of images, whereas previous methods rely on either a single image or the entire dataset.}
    \label{fig:intro}
\end{figure}

The advancement of vision-language pre-training~\cite{radford2021learning} and their strong zero-shot generalization abilities have led many recent studies to incorporate them into zero-shot anomaly detection. 
%
In particular, WinCLIP~\cite{jeong2023winclip} and APRIL-GAN~\cite{april-gan} initially employed CLIP~\cite{radford2021learning} in anomaly detection, leveraging the similarity between image and text features to estimate anomaly probability. 
As CLIP~\cite{radford2021learning} was developed for classification tasks, some works~\cite{anomalyclip, clip-ad, filo} observe that the original $q-k$ attention disrupts local semantics and limits fine-grained anomaly segmentation, where $q$, $k$, $v$ represent the $query$, $key$, $vlaue$ embeddings.
They introduced $v-v$ attention to strengthen local semantic representation.
Although improvements in the self-attention mechanism have led to performance gains, local semantic representation remains insufficient, and modifications across multiple layers have somewhat compromised the model's generalization ability. In addition to improvements on the visual level, studies like AnomalyCLIP~\cite{anomalyclip}, AdaCLIP~\cite{adaclip} and Filo~\cite{filo} have also designed learnable text prompts, aiming to enable the model to understand the concepts of `normal' and `abnormal'. 
Additionally, MuSc~\cite{musc} finds that over 95\% of the pixels in the test set of MVTec AD dataset~\cite{bergmann2019mvtec} and VisA dataset~\cite{zou2022spot} are normal, so it utilizes the entire test dataset for feature matching in anomaly detection. 
However, There are certain problems associated with the aforementioned methods. 
First, in practical production environments, testing is usually conducted on batches of samples, while applying the entire test dataset in real-world industrial scenarios is impractical and requires extensive computational resources. Moreover, the posterior optimization in MuSc~\cite{musc} requires prior knowledge of the test dataset distribution before testing.
Second, most of ZSAD methods involve fine-tuning, a setup that may risk data leakage. 
%

To tackle the abovementioned issues, we propose training-free FiSeCLIP, which combines mainstream zero-shot anomaly detection principles with the idea of feature matching.
\textbf{\textit{During testing, we process images in batches, leveraging cross-reference cues within the batch to extract limited reference information, thus requiring simultaneous input of all batch images to the model.}} 
As shownin Fig\ref{fig:intro}, we will apply a strategy similar to MuSc~\cite{musc} and PatchCore~\cite{patchcore}, denoted FiCLIP-AD. 
As input images are unlabeled test images, this can lead to noise interference. In contrast to MuSc\cite{musc}, we propose a filtering strategy to filter out anomalous features from the unknown label reference information.
The \textit{\textbf{filtering mask}} is dynamically constructed via cross-modal similarity alignment between CLIP's visual embeddings and their corresponding textual embeddings, named SeCLIP-AD, to optimize feature-matching outcomes. 
Specifically, we find that intermediate-layer attention provides stronger local semantic representation, while the residual connection and the feed-forward network can degrade fine-grained segmentation performance~\cite{clearclip}. To retain CLIP's generalization ability, we replace only the final layer's attention with intermediate-layer attention and remove the feed-forward network. 
%
%
%
Furthermore, by integrating the results of FiCLIP-AD and SeCLIP-AD, the fused outcomes enable iterative refinement of the \textit{\textbf{filtering mask}}. This mutual optimization loop enhances the accuracy of FiCLIP-AD’s anomaly score map through continuous feedback, where the refined \textit{\textbf{filtering mask}} guides more precise feature matching and noise suppression. 
In summary, our contributions can be outlined as follows:
\begin{itemize}
  \item We propose a novel method, FiSeCLIP, which integrates feature matching with mainstream ZSAD techniques. By leveraging unlabeled test images within the same batch as mutual references, FiSeCLIP introduces a mutual-filtering strategy to dynamically filter out anomalous features and mitigate noise interference during feature matching.
  \item We begin by using textual information to perform initial filtering of noisy features, followed by further refinement with feature matching results. We propose enhancing CLIP’s semantic coherence to improve the alignment of fine-grained visual and textual features.
  \item We perform comprehensive experiments, achieving state-of-the-art performance on widely used benchmarks, e.g., we achieved improvements of +4.6\%$\uparrow$/+5.7\%$\uparrow$ and +6.1\%$\uparrow$/+4.8\%$\uparrow$ in AU-ROC/$F_1$-max for anomaly segmentation and classification, respectively.
\end{itemize}

\section{Related Work} 
\label{sec:related_work}

\subsection{Industrial anomaly detection}


Most industrial anomaly detection works can be classified into three categories: 1) reconstruction-based methods; 2) synthesizing-based methods and 3) embedding-based methods. \textbf{Reconstruction-based methods}~\cite{ocr, CAVGA,anomalydiffusion,wyatt2022anoddpm,mambaad,vitad}posit that anomalous regions are difficult to reconstruct correctly. Thus, reconstruction error serves as an indicator for locating anomalous areas. \textbf{Synthesizing-based methods}~\cite{draem, cutpaste, realnet, unified} generate artificial anomalies to provide supervision signals for training. \textbf{Embedding-based methods}~\cite{patchcore, padim, pni,REB,cflow,DFD} extract features using a pre-trained network, then establish a memory bank to detect anomalous patches through feature matching. However, these methods require a large number of samples and have limited generalization capability. 

\subsection{Vision-language models for ZSAD}
\noindent\textbf{Vision-language pre-training models.} Vision-language pre-training has emerged as a powerful approach for visual representation learning, with CLIP ~\cite{radford2021learning} standing out due to its impressive generalization capabilities. 
Pretrained on a vast dataset of images sourced from the web, CLIP aligns images and natural language through two separate encoders, typically based on architectures like ResNet ~\cite{he2016deep}, ViT ~\cite{dosovitskiy2020image}, or their enhanced versions. 
This design allows CLIP to be easily adapted to various downstream classification tasks using prompts.
Despite being originally intended for classification, CLIP has been adapted for fine-grained segmentation~\cite{lan2024proxyclip, shao2025explore, clearclip, clip-surgery,sclip} without any training.
These works improve local feature awareness with a modified attention layer, addressing the dominance of global patches and boosting segmentation accuracy and semantic coherence. The anomaly detection task is closely related to fine-grained segmentation task, and similar research ideas have been explored and extended to anomaly detection in later studies.

\noindent \textbf{Zero-shot anomaly detection.} 
WinCLIP~\cite{jeong2023winclip} pioneers the application of CLIP in zero-shot anomaly detection. Subsequently, APRIL-GAN~\cite{april-gan} improves upon it by incorporating a linear layer for fine-tuning. Some works~\cite{anomalyclip, clip-ad, filo} modify the $q-k$ attention to $v-v$ attention to enhance CLIP's local semantics. Moreover, several methods~\cite{adaclip, anomalyclip, filo, vcp-clip} employ learnable text prompts to boost zero-shot generalization, with the goal of enabling the model to autonomously identify the concepts of `anomaly' and `normal'. However, the limited alignment of CLIP's visual and textual features in fine-grained tasks severely affects ZSAD performance. Furthermore, the need for fine-tuning in many prior approaches raises the risk of data leakage. Recently, MuSc~\cite{musc} demonstrates outstanding performance by leveraging the entire test set with the idea of feature-matching. Nevertheless, using the entire dataset is impractical for real-world anomaly detection and necessitates substantial memory resources. 
\section{Method} \label{sec:method}


This section presents our proposed FiSeCLIP, divided into two components: FiCLIP-AD and SeCLIP-AD. 
FiCLIP-AD uses CLIP to extract features from unlabeled images within a batch, with each image as a reference for the others. 
However, unlabeled references may introduce noise interference. To address this, we propose SeCLIP-AD to first generate a prior filtering mask that suppresses anomalous features, ensuring cleaner reference information for subsequent feature matching.
SeCLIP-AD investigates a training-free strategy that harnesses CLIP's intrinsic capabilities to enhance local semantic self-correlation among patches. The anomaly score $A_{Se}$ is obtained by calculating the similarity between visual and text features. 
 We seamlessly integrate FiCLIP-AD and SeCLIP-AD predictions and iteratively feed them back to FiCLIP-AD, significantly enhancing anomaly score map accuracy while suppressing false positives.
%

\subsection{Task setting.}
In this study, we perform simultaneous testing on images within a batch $D = \{I_u\}, u = 1, ..., B$, using them as mutual references, a method we refer to as batch zero-shot anomaly detection. In contrast, mainstream zero-shot anomaly detection\cite{jeong2023winclip,april-gan, anomalyclip, adaclip} requires only a single image $D = \{I_u\}, u = 1$ for inference. The recent MuSc\cite{musc} requires the entire test dataset $D = \{I_u\}, u = 1, ..., n$ to be processed simultaneously for inference. MuSc-2\cite{musc} and Dual Image Enhanced CLIP\cite{Dual-Image} adopt two images from the test set for evaluation, while ACR utilizes 16 images from the test set. The remaining methods\cite{jeong2023winclip, april-gan, anomalyclip, adaclip} all take a single image as input and output the results. For simplicity, we will use two images ($I_u$ and $I_v$) as examples to introduce our method in the following sections.

\subsection{SeCLIP-AD}
\label{seclip}
\begin{figure*}[h]
    \centering
    \includegraphics[width=0.85\linewidth, trim=90 85 90 0,clip]{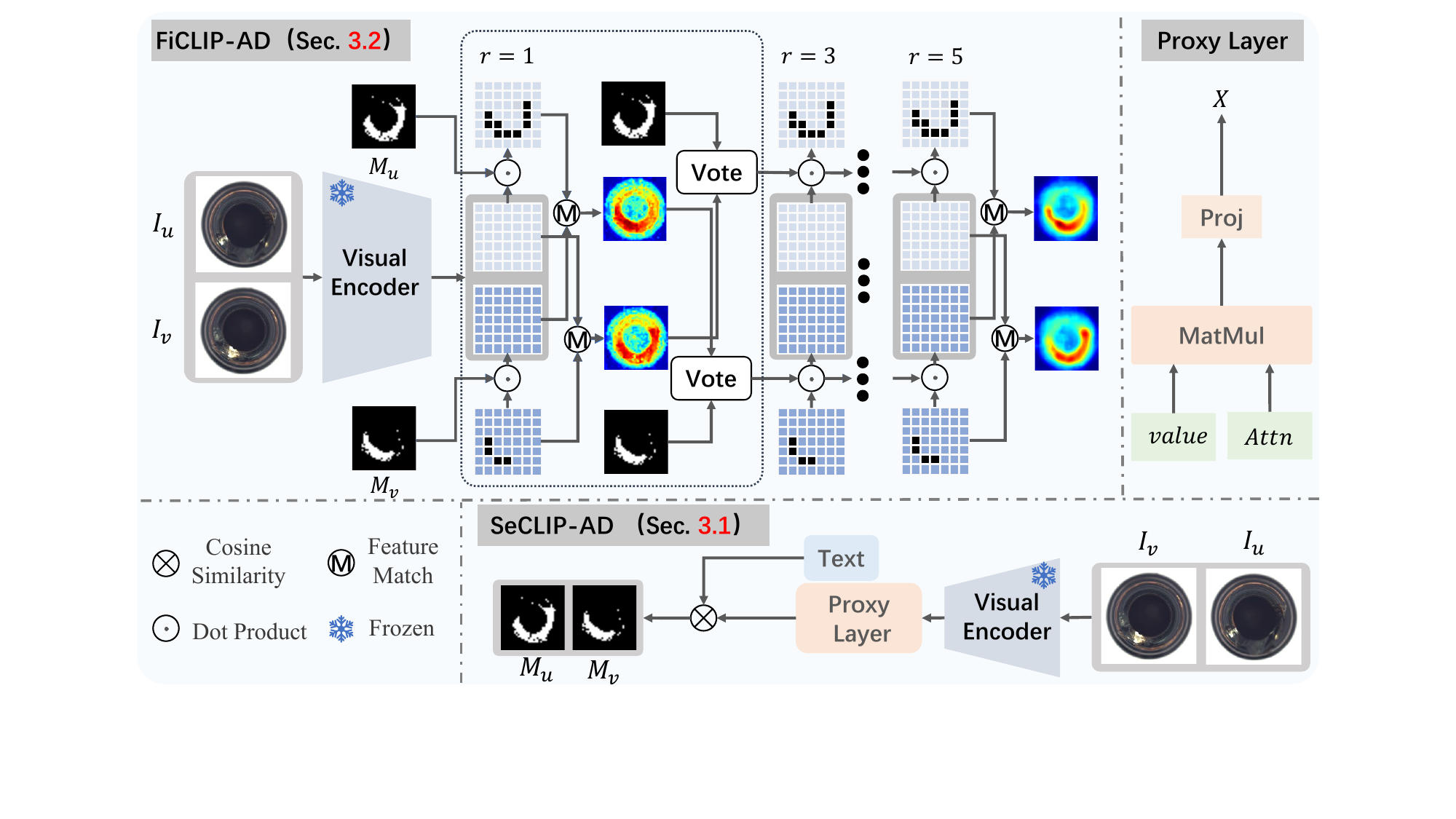}
    \caption{\textbf{Overview of our FiSeCLIP}. It consists of two parts: SeCLIP-AD (Sec. \ref{seclip}) and FiCLIP-AD (Sec. \ref{ficlip}). 1) For two images in the same batch, SeCLIP-AD calculates the similarity between text and visual features to generate the corresponding anomaly score and mask, which are then used by the FiCLIP-AD module. The Proxy Layer is used to restore semantic correlation among patches, with attention weights derived from the intermediate layer. 2) For FiCLIP-AD, two images serve as inputs for feature extraction and aggregation with varying neighbor counts ($r=1, 3, 5$). The anomaly mask from the previous step and the mask generated by SeCLIP-AD are refined through mutual voting, and the final mask is used to filter noisy features for feature matching. Different shades of blue represent patch tokens from different images.}
    \label{fig:method}
\end{figure*}

\subsubsection{Semantic correlation recovering }

CLIP is trained for classification tasks with more attention to global semantic information. 
Meanwhile, anomaly classification and segmentation are required to understand local semantics. 
Thus, we need to recover its semantic correlation for the local patterns.

A ViT-based CLIP vision encoder comprises a series of attention blocks. 
Each block yield the feature representation $X_i \in \mathbb{R}^{B \times (HW+1) \times D}$, where $i$ represents the layer index, $B$ is the batch size, $D$ is the dimension, $HW$ is the number of local patch tokens, and the other one is $[CLS]$ token. For simplicity, a self-attention-based ViT encoder is described as follows:
\begin{equation}
    [q_i, k_i, v_i] = \text{Proj}_{qkv}(\text{LN}(X_{i-1})), 
\end{equation}
\begin{equation}
    \text{Attn}_i = \text{Softmax}(\frac{q_i^T k_i}{\sqrt{d_k}}),
\end{equation}
\begin{equation}
    X_i = X_{i-1} + X_{attn} = X_{i-1} + \text{Proj}(\text{Attn}_i \cdot v_i),
\end{equation}
\begin{equation}
    X_i = X_i + \text{FFN}(\text{LN}(X_i),
\end{equation}
where Proj denotes a projection layer, LN denotes layer normalization, and FFN represents a feed-forward network. $q$, $k$, and $v$ represent the query, key, and value embeddings, respectively. $i$ is the layer index, and $i-1$ denotes the output from the previous layer.

\begin{figure}
    \centering
    \includegraphics[width=1\linewidth]{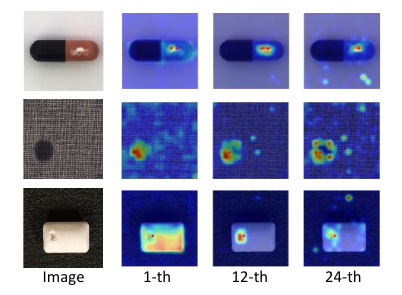}
    \caption{\textbf{Attention maps from different layers of CLIP}, with red dots indicating the selected patch positions in the image, which shows 12-th attention map has the best local semantics.}
    \label{fig:attn}
\end{figure}

To retain CLIP's generalization capacity, we restrict changes to the last layer. 
Visualizing the CLIP's attention map, as shown in Fig.\ref{fig:attn}, it can be found that the middle layer's attention map has better local semantics. 
As a result, the intermediate layer's attention $\text{Attn}_{inter}$ is substituted for that of the final layer attention map $\text{Attn}_{-1}$:
\begin{equation}
    \text{Attn}_{-1} = \text{Attn}_{inter}, X_{attn} = \text{Proj}(\text{Attn}_{inter} \cdot v_{-1}).
\end{equation}
What's more, ClearCLIP\cite{clearclip} finds residual connection significantly degrade performance on dense segmentation tasks, with the feed-forward module having a negligible impact. 
In this paper, we adopt a similar approach. Specifically, the output $X_{final} \in \mathbb{R}^{B \times (HW+1) \times D }$ of the final block layer can be defined as $X_{final} = X_{attn}$.

\subsubsection{CLIP for Zero-shot Anomaly Detection}
To leverage CLIP's multi-modal capability for zero-shot anomaly detection, a naive idea is to compute the similarity between text features and visual features.

Anomaly detection involves the concepts of normal and abnormal, so the text prompts can be simply designed as `a [d] photo of [s] [c]', where [s] represents the concepts of normality and abnormality, such as `normal' and `abnormal,' or `flawless' and `damaged.' [d] offers additional information about the image, such as `rotated' or `cropped,' and [c] indicates the image's category. 
Two types of text prompts, normal and abnormal, are generated. We utilize CLIP’s text encoder to extract text features, and the final text features $F_t \in \mathbb{R}^{2 \times C }$ can be obtained by averaging the features of normal and abnormal prompts separately, where $C$ denotes the dimension of the features.

For anomaly classification, the visual global features $F_c \in \mathbb{R}^{B \times C}$ of original CLIP is utilized, and the probability of an object $Cls_{prob} \in \mathbb{R}^{B \times 2}$ being classified as normal or abnormal can be expressed as:
\begin{equation}
    CLS_{prob} = \text{Softmax}(F_c \cdot F_t).
\end{equation}
The anomaly classification score $CLS_{score} \in \mathbb{R}^B$ refers to the probability of the category being abnormal.

For anomaly segmentation, we remove the [CLS] token and map the dimension to $C$, ultimately obtaining the local patch features $F_s \in \mathbb{R}^{B \times HW \times C}$. 
The anomaly segmentation probability $SEG_{prob}\in\mathbb{R}^{B \times H \times W \times 2}$ can be described as:
\begin{equation}
\label{eq:seg_prob}
    SEG_{prob} = \text{Softmax}(F_s \cdot F_t),
\end{equation}
and the anomaly segmentation score $Seg_{score} \in \mathbb{R}^{B \times HW}$ refers to the probability of the patch being abnormal.

\subsection{FiCLIP-AD}
\label{ficlip}
\subsubsection{Feature extraction}
\label{subsubsection:feature extraction}
We use the original CLIP to simultaneously extract the features of $B$ unlabeled test images $D = \{I_u\}, u=1,...,B$. 
For the ViT-based CLIP encoder with $L$ layers, the multi-stage patch tokens $F_u^i \in \mathbb{R}^{HW \times D}$ are employed, where $i \in \{0, 1, ...,L\}$ indicates the layer of CLIP visual encoder. 
By aggregating neighboring features, we enhance the detection accuracy for anomalies across various sizes. Specifically, the patch tokens $F_u^i \in \mathbb{R}^{HW \times D}$ can be reshaped into $H \times W \times D$. 
Average pooling is employed to aggregate the patch token features in an $r \times r$ neighborhood of the current position to get the aggregated token $F_u^{i,r} \in \mathbb{R}^{H \times W \times D}$ following \cite{patchcore, musc}, and reshape it to $F_u^{i,r} \in \mathbb{R}^{HW \times D}$. 

\subsubsection{Noisy features filtering}
\begin{figure}
        \centering
        \includegraphics[width=1\linewidth, trim=80 80 80 0, clip]{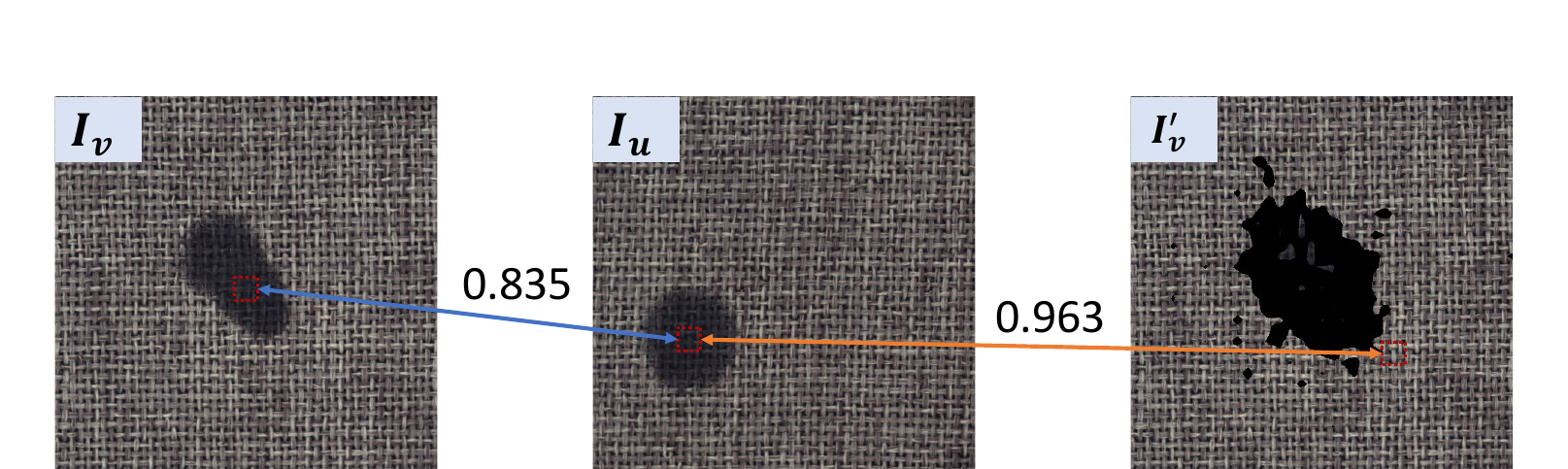}
        \caption{An example of mutual-filtering. $I_u$ denotes the inference image, $I_v$ represents the reference image without filtering, and $I_v'$ represents the reference image with filtering.}
        \label{fig:filter anomaly features}
\end{figure}

For the $B$ images in the same batch size, the corresponding features are extracted using the method outlined in Sec.\ref{subsubsection:feature extraction}. 
Calculate the distance between each patch token of the visual features $F_u^{i,r} \in \mathbb{R}^{HW \times D}$ and $F_v^{i,r} \in \mathbb{R}^{HW \times D}$, $v \neq u$, taking the minimum distance as the anomaly score $a_{u,v}^{i,r} \in \mathbb{R}^{HW}$:
\begin{equation}
    a_{u,v}^{i,r} = \text{min}|| F_u^{i,r} - F_v^{i,r}||_2,
\end{equation}
In applying feature-match methods\cite{patchcore, santos2023optimizing} for anomaly detection, we aim for the reference images to contain no patches similar to the anomalous patches in the inference image. 
However, when applying the idea to zero-shot anomaly detection, the uncertainty of whether the reference features are normal can introduce some interference. 
As shown in Fig.\ref{fig:filter anomaly features}, the similarity between anomalous patches in the reference and inference images results in a decreased anomaly score for the inference image, which may lead to a higher rate of false negatives in anomaly detection. 
Therefore, we propose filtering out the anomalous features in the reference features $F_v^{i,r}$ and using the filtered features as references to calculate the anomaly score. 
The initial anomaly mask $M \in \mathbb{R}^{HW}$ can be obtained in Eq.\ref{eq:seg_prob}:
\begin{equation}
\label{lamdda}
    M =   
    \begin{cases}
    True, & \text{if } P_a > \lambda \cdot P_n\\
    False, & \text{otherwise}
    \end{cases}
\end{equation}
where $P_a$ and $P_n$ represent the abnormal and normal probabilities in Eq.\ref{eq:seg_prob}. Consequently, the `normal' features can be defined as:
\begin{equation}
    \hat{F}_v^{i,r} =  F_v^{i,r}[\bar{M}],
\end{equation}
where $\hat{F}_v^{i,r} \in \mathbb{R}^{N' \times D}$, and $N'$ denotes the length of normal patches. The anomaly score is further described as follows:
\begin{equation}
\label{eq:anomaly score}
    \hat{a}_{u,v}^{i,r} = \text{min}|| F_u^{i,r} - \hat{F}_v^{i,r}||_2,
\end{equation}

\subsubsection{Mutual-filtering mask}
In Sec.\ref{subsubsection:feature extraction}, the features $F_u^{i,r}$ from multiple stages and various aggregation types are extracted. Moreover, Eq.\ref{eq:anomaly score} allows for obtaining multiple anomaly scores. 
Therefore, we propose using the anomaly score to refine the anomaly mask $M$. Specifically, the four stages patch tokens are utilized to generate the average anomaly score:
\begin{equation}
\begin{aligned}
    \hat{a}_v^r &= \frac{1}{m}\sum_{i\in \{0, 1, ...,L\}} \hat{a}_{v,u}^{i,r},\\
    \hat{a}_u^r &= \frac{1}{m}\sum_{i\in \{0, 1, ...,L\}} \hat{a}_{u,v}^{i,r},
\end{aligned}
\end{equation}
where $m$ indicates the number of stages employed. The intermediate mask is defined as follows:
\begin{equation}
\label{mu}
    M_{inter} = 
    \begin{cases}
    True, & \text{if } \hat{a}^r > \mu\\
    False, & \text{otherwise}
    \end{cases}
\end{equation}
where $\mu$ is a hyperparameter with a default value of 0.57, and $\hat{a}^r$ denotes the anomaly score of an image. 
Then apply collaborative voting with intermediate mask $M_{inter}$ to refine $M$. The detailed process is illustrated in Alg.\ref{alg:mutual-filtering}. 

\begin{algorithm}[ht]
\SetAlgoLined
\KwInput{Starting masks $M_u$ and $M_v$, inference features $F_u^{i,r}$, reference features $F_v^{i,r}$, stage index $i$, aggregation neighborhood $r$, initial anomaly score $\hat{a}$.}
\KwOutput{Anomaly score $\hat{a}_u, \hat{a}_v$.}
\textbf{Algorithm:}\\
$\hat{a}_u, \hat{a}_v \leftarrow empty$\\
 \For{$r\in {1, 3, 5}$}{
        \For{$i \in {6, 12, 18, 24}$}{
        $\hat{F}_u^{i,r} \leftarrow F_u^{i,r}[\bar{M_u}]$, $\hat{F}_v^{i,r} \leftarrow F_v^{i,r}[\bar{M_v}]$\\
        
        $a_{u}^{i,r} \leftarrow \text{min}|| F_u^{i,r} - \hat{F}_v^{i,r}||_2$\\
        $a_{v}^{i,r} \leftarrow \text{min}|| F_v^{i,r} - \hat{F}_u^{i,r}||_2$\\
        }
 	$\hat{a}_u^r \leftarrow \frac{1}{4}\sum \hat{a}_{u}^{i,r}$, $\hat{a}_v^r \leftarrow \frac{1}{4}\sum \hat{a}_{v}^{i,r}$\\  
  
        $\hat{a}_u \leftarrow \text{mean}(\hat{a} + \hat{a}_u^r)$, $\hat{a}_v \leftarrow \text{mean}(\hat{a}_v + \hat{a}_v^r)$\\  
        
        $M_u^{inter} \leftarrow \text{binary}(\hat{a}_u^r)$, $M_v^{inter} \leftarrow \text{binary}(\hat{a}_v^r)$\\

        $M_u \leftarrow \text{vote}(M_u, M_u^{inter})$\\
        $M_v \leftarrow \text{vote}(M_v, M_v^{inter})$\\
 }
\caption{Noise mutual filtering}
\label{alg:mutual-filtering}
\end{algorithm}
\section{Experiments} 
\label{sec:exp}


\begin{table*}
  \centering
  \belowrulesep=0pt
  \aboverulesep=0pt
  \renewcommand{\arraystretch}{1.2}
  \caption{\textbf{Comparison with some state-of-the-art on zero-shot and 1-shot methods.} Bold values indicate the best results, while underlined values represent the suboptimal results. MuSc-2 denotes inference with 2 images.}
  \label{tab:comparison with sotas}
  \resizebox{0.9\linewidth}{!}{%
    \begin{tabular}{c|cccccccccc}
\toprule[1.5pt]
\multirow{2}{*}{Datasets} & \multirow{2}{*}{Methods}    & \multirow{2}{*}{Train} & \multirow{2}{*}{Setting} & \multicolumn{4}{c}{Segmentation} & \multicolumn{3}{c}{Classification} \\
\cmidrule(r){5-8} \cmidrule(l){9-11}
                          &  &   &  & AU-ROC   & $F_1$-max     & AP    & AU-PRO& AU-ROC & $F_1$-max & AP        \\
\midrule
\multirow{10}{*}{MVTec}   & WinCLIP\cite{jeong2023winclip}  & \xmarkg  & 0-shot & 85.1           & 31.7           & 18.2           & 64.6           & \underline{91.8}           & 92.9           & \underline{96.5}           \\
                          & APRIL-GAN\cite{april-gan}       & \cmark   & 0-shot & 87.6           & 43.3           & 40.8           & 44.0           & 86.1           & 90.4           & 93.5           \\
                          & AnomalyCLIP\cite{anomalyclip}   & \cmark   & 0-shot & 91.1           & -              & -              & 81.4           & 91.5           & -              & 96.2           \\
                          & AdaCLIP\cite{adaclip}           & \cmark   & 0-shot & 88.7           & 43.4           & -              & -              & 89.2           & 90.6           & -              \\
                          & ACR\cite{acr}  & \cmark & 0-shot & 92.5 & - & - & - & 85.8 &  - & - \\ 
                          & \cite{Dual-Image} & \cmark & 0-shot & 92.8 & 42.5 & - & 84.0 & 93.2 & 94.1 & 96.7\\
                          & MuSc-2\cite{musc}               & \xmarkg  & 0-shot & \underline{92.6}           & \underline{46.8}           & \underline{42.8}           & \underline{86.4}           & 89.7           & \underline{93.0}           & 95.2           \\
                           & \cellcolor{blue!10}\textbf{Ours}                            & \cellcolor{blue!10}\xmarkg  & \cellcolor{blue!10}0-shot &\cellcolor{blue!10} \textbf{93.3}  &\cellcolor{blue!10} \textbf{49.1}  & \cellcolor{blue!10}\textbf{46.5}  &\cellcolor{blue!10} \textbf{89.6}  &\cellcolor{blue!10} \textbf{95.3}  &\cellcolor{blue!10} \textbf{95.4}  &\cellcolor{blue!10} \textbf{98.3}  \\
\cmidrule{2-11} 
                          & PatchCore\cite{patchcore}       & \xmarkg  & 1-shot & 93.3           & 53.0           & -              & 82.3           & 86.3           & 92.0           & 93.8           \\
                          & WinCLIP\cite{jeong2023winclip}  & \xmarkg  & 1-shot & \textbf{95.2}  & \textbf{55.9}  & -              & 87.1           & \underline{93.1}           & \underline{93.7}           & \underline{96.5}           \\
                          & APRIL-GAN\cite{april-gan}       & \cmark   & 1-shot & \underline{95.1}           & \underline{54.2}           & \textbf{51.8}  & \textbf{90.6}  & 92.0           & 92.4           & 95.8           \\
                          & \cellcolor{blue!10}\textbf{Ours}                            & \cellcolor{blue!10}\xmarkg  &\cellcolor{blue!10} 0-shot &\cellcolor{blue!10} 93.3           & \cellcolor{blue!10}49.1           & \cellcolor{blue!10}\underline{46.5}           &\cellcolor{blue!10} \underline{89.6}           &\cellcolor{blue!10} \textbf{95.3}  & \cellcolor{blue!10}\textbf{95.4}  & \cellcolor{blue!10}\textbf{98.3}  \\
\midrule
\multirow{10}{*}{VisA}    & WinCLIP\cite{jeong2023winclip}  & \xmarkg  & 0-shot & 79.6           & 14.8           & 5.4            & 56.8           & 78.1           & 79.0           & 81.2           \\
                          & APRIL-GAN\cite{april-gan}       & \cmark   & 0-shot & 94.2           & 32.3           & 25.7           & 86.8           & 78.0           & 78.7           & 81.4           \\
                          & AnomalyCLIP\cite{anomalyclip}   & \cmark   & 0-shot & 95.5           & -              & -              & \underline{87.0}           & 82.1           & -              & \underline{85.4}           \\
                          & AdaCLIP                         & \cmark   & 0-shot & 95.5           & \textbf{37.7}  & -              & -              & \textbf{85.8}  & \underline{83.1}           & -              \\
                         & \cite{Dual-Image} & \cmark & 0-shot & 94.2 & 24.1 & - & 79.7 & 82.9 & 80.9 & 84.7\\
                          & MuSc-2\cite{musc}               & \xmarkg  & 0-shot & \underline{95.6}           & 32.6           & \underline{26.1}           & 80.5           & 74.0           & 77.5           & 78.1           \\
                          & \cellcolor{blue!10}\textbf{Ours}                            & \cellcolor{blue!10}\xmarkg  &\cellcolor{blue!10} 0-shot &\cellcolor{blue!10} \textbf{97.4}  & \cellcolor{blue!10}\underline{35.8}           & \cellcolor{blue!10}\textbf{30.8}  & \cellcolor{blue!10}\textbf{87.6}  & \cellcolor{blue!10}\underline{85.5}           & \cellcolor{blue!10}\textbf{83.5}  & \cellcolor{blue!10}\textbf{88.2}  \\
\cmidrule{2-11} 
                          & PatchCore\cite{patchcore}       & \xmarkg  & 1-shot & 95.4           & 38.0           & -              & 80.5           & 79.9           & 81.7           & 82.8           \\
                          & WinCLIP\cite{jeong2023winclip}  & \xmarkg  & 1-shot & \underline{96.4}           & \textbf{41.3}  & -              & 85.1           & 83.8           & 83.1           & 85.1           \\
                          & APRIL-GAN\cite{april-gan}       & \cmark   & 1-shot & 96.0           & \underline{38.5}           & \textbf{30.9}  & \textbf{90.0}  & \textbf{91.2}  & \textbf{86.9}  & \textbf{93.3}  \\
                          & \cellcolor{blue!10}\textbf{Ours}                            & \cellcolor{blue!10}\xmarkg  &\cellcolor{blue!10} 0-shot & \cellcolor{blue!10}\textbf{97.4}  & \cellcolor{blue!10}35.8           & \cellcolor{blue!10}\underline{30.8}           &\cellcolor{blue!10} \underline{87.6}           & \cellcolor{blue!10}\underline{85.5}           & \cellcolor{blue!10}\underline{83.5}           & \cellcolor{blue!10}\underline{88.2}           \\
\bottomrule[1.5pt]
\end{tabular}

  }
  \vspace{-0.5em}
\end{table*}

\subsection{Experimental Setups}
\noindent\textbf{Datasets.}
Our experiments are primarily conducted on the MVTec AD \cite{bergmann2019mvtec}, VisA\cite{zou2022spot} and BTAD\cite{btad} datasets. The MVTec dataset\cite{bergmann2019mvtec} comprises 15 categories and a total of 5,354 images, of which 3,629 are for training, and 1,725 are for testing, with resolutions varying between $700 \times 700$ and $1,024 \times 1,024$ pixels. The VisA dataset\cite{zou2022spot} provides 12 categories with a total of 10,821 images, including 9,621 normal images and 1,200 anomalous images, with a resolution of $1,500 \times 1,000$ pixels. 

\noindent\textbf{Evaluation Metrics.}
Following previous works\cite{jeong2023winclip, april-gan, musc, mambaad}, we employ the Area Under the Receiver Operating Characteristic Curve (AU-ROC), F1-score at optimal threshold ($F_1$-max)\cite{zou2022spot} and Average Prevision (AP)\cite{draem} for both anomaly classification and anomaly segmentation. Per-Region Overlap (AU-PRO)\cite{pro} is also utilized for anomaly segmentation. 

\noindent\textbf{Implementation Details.}
All test images have been resized to a resolution of $518 \times 518$. We employ the pre-trained CLIP model with ViT-L/14-336\cite{alexey2020image, radford2021learning} as our backbone, keeping all parameters frozen throughout the experiments. The model consists of 24 layers, organized into 4 stages with 6 layers each. Patch tokens are extracted from the outputs of 6-th, 12-th, 18-th, and 24-th layers. The CLS token used at the image level is derived from the original CLIP output. In addition, we only alter the final layer of CLIP in Sec.\ref{seclip}. All experiments are conducted on a single RTX 3090 GPU.

\begin{figure*}[h]
    \centering
    \includegraphics[width=1\linewidth, trim=80 180 80 0,clip]{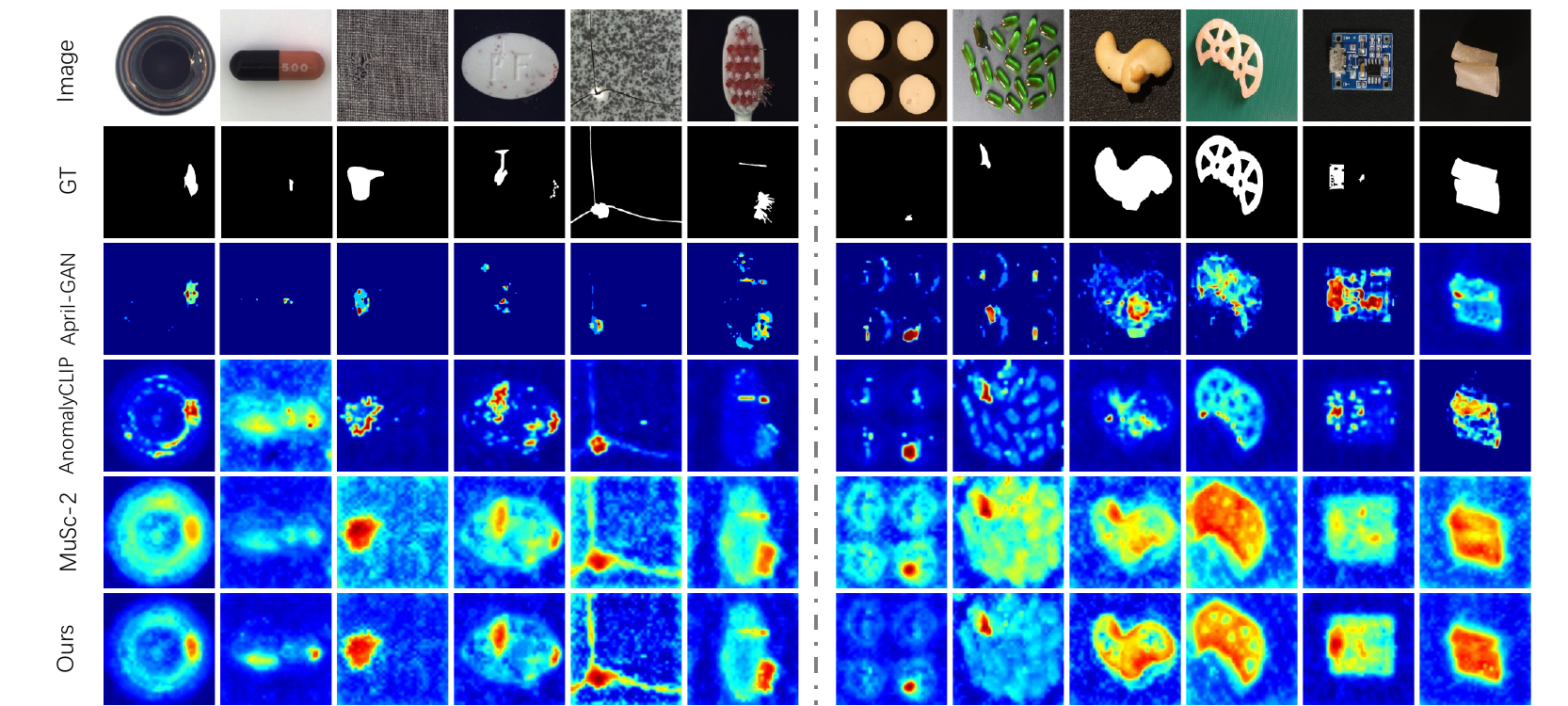}
    \caption{\textbf{Visualization results of anomaly localization for each class on MVTec AD (left) dataset and VisA dataset (right).}}
    \label{fig:visual}
\end{figure*}

\subsection{Main Results}
We compare our method with SoTA methods on anomaly classification and segmentation metrics for zero-shot and 1-shot anomaly detection. 

\noindent\textbf{Quantitative Results.}
According to Tab.\ref{tab:comparison with sotas}, on MVTec AD dataset\cite{bergmann2019mvtec} we demonstrate a substantial enhancement compared to the training-free WinCLIP, which uses manually designed text prompts. Our method achieved an 8.2\%$\uparrow$ and 3.5\%$\uparrow$ increase on AU-ROC for anomaly segmentation and classification, respectively, and over a 20\%$\uparrow$ improvement in AP and AU-PRO for anomaly segmentation. Additionally, our method outperforms our baseline, MuSc\cite{musc}, with an 0.7\%$\uparrow$ increase in AU-ROC for anomaly segmentation and a 5.6\%$\uparrow$ increase for anomaly classification. Compared to AnomalyCLIP, which is fine-tuned on an auxiliary dataset, our method achieves an additional 2.2\%$\uparrow$ in AU-ROC for segmentation and 3.8\%$\uparrow$ for classification.
Furthermore, compared to 1-shot methods (using one normal image as a reference), our approach still demonstrates superior performance in anomaly classification. For instance, we achieve improvements of 2.2\%$\uparrow$, 1.7\%$\uparrow$, and 1.8\%$\uparrow$ on AU-ROC, $F_1$-max, and AP for anomaly classification compared to WinCLIP\cite{jeong2023winclip}. On VisA dataset\cite{zou2022spot}, we continue to achieve SOTA performance, especially in anomaly segmentation, where we significantly outperform previous methods and even exceed the segmentation performance of 1-shot anomaly segmentation approaches. 

\noindent\textbf{Qualitative Results.}
As shown in the Fig.\ref{fig:visual}, we visualized anomaly segmentation on a subset of data from the MVTec AD\cite{bergmann2019mvtec} and VisA\cite{zou2022spot} datasets. It can be observed that our method accurately localizes anomalies in most objects, demonstrating the effectiveness of our approach. Compared with the previous methods~\cite{anomalyclip, april-gan, musc}, our method has better anomaly segmentation results.
\begin{table}
  \centering
  \belowrulesep=0pt
  \aboverulesep=0pt
  \renewcommand{\arraystretch}{1.2}
  \caption{Experiments on different noise filtering methods. Bold values indicate the best results}
  \label{tab:filter_methods}
  \resizebox{\linewidth}{!}{%
    \begin{tabular}{@{}c|ccccccc@{}}
    \toprule[1.5pt]
    \multirow{2}{*}{Method} & \multicolumn{4}{c}{Segmentation} & \multicolumn{3}{c}{Classification}\\
    \cmidrule(r){2-5}
    \cmidrule(r){6-8}
    ~ & AU-ROC& $F_1$-max& AP & AU-PRO& AU-ROC & $F_1$-max & AP \\
    \midrule
    w/o filtering&92.8 & 46.6 & 42.7 & 86.5 & 89.4 & 92.7 & 94.9 \\
    $M$&93.1 & 47.9 & 44.9 & 88.2 & 92.8 & 93.3 & 96.8 \\
    $M^{r-,i}$&92.8 & 46.7 & 42.7 & 86.6 & 89.5 & 92.8 & 94.9 \\
    $M^{i,r}$&93.1 & 48.1 & 45.2 & 87.8 & 91.7 & 93.2 & 96.9 \\
    $M^{r*i}$&93.1 & 48.2 & 45.4 & 88.2 & 93.0 & 93.8 & 97.0 \\
    \cellcolor{blue!10}$M^{r,i}$& \cellcolor{blue!10}\textbf{93.2} & \cellcolor{blue!10}\textbf{48.3} &\cellcolor{blue!10} \textbf{45.5} & \cellcolor{blue!10}\textbf{88.5} &\cellcolor{blue!10} \textbf{93.5} & \cellcolor{blue!10}\textbf{94.7} &\cellcolor{blue!10} \textbf{97.3} \\
    \bottomrule[1.5pt]
\end{tabular}
  }
  \vspace{-0.5em}
\end{table}

\begin{table*}[ht]
\centering
\begin{minipage}[t]{0.47\linewidth}
  \centering
  \caption{Comparative study of the anomaly classification and segmentation performance on MVTec AD with various $\lambda$ settings. Bold values indicate the best results.}
  \resizebox{\linewidth}{!}{%
    \begin{tabular}{@{}c|ccccccc@{}}
    \toprule[1.5pt]
    \multirow{2}{*}{$\lambda$} & \multicolumn{4}{c}{Segmentation} & \multicolumn{3}{c}{Classification}\\
    \cmidrule(r){2-5}
    \cmidrule(r){6-8}
    ~ & AU-ROC& $F_1$-max& AP & AU-PRO& AU-ROC & $F_1$-max & AP \\
    \midrule
    0.95 & 93.0 & 47.9 & 44.8 & 88.3 & \textbf{93.6} & 94.6 & 97.3 \\
    1.00  & 93.1 & 48.0 & 44.9 & 88.3 & 93.5 & 94.6 & 97.3 \\
    1.05 & 93.1 & 48.0 & 44.9 & 88.3 & 93.5 & 94.6 & 97.3 \\
    \cellcolor{blue!10}1.10  & \cellcolor{blue!10}\textbf{93.2} & \cellcolor{blue!10}\textbf{48.1} &\cellcolor{blue!10}\textbf{ 45.0} & \cellcolor{blue!10}\textbf{88.5} & \cellcolor{blue!10}93.5 & \cellcolor{blue!10}\textbf{94.7} & \cellcolor{blue!10}97.3 \\
    1.20  & 93.1 & 48.1 & 44.9 & 88.3 & 93.4 & 94.3 & 97.3 \\
    \bottomrule[1.5pt]
\end{tabular}
  }
  \label{table:lambda}
\end{minipage}
\hspace{0.02\textwidth}
\begin{minipage}[t]{0.47\linewidth}
  \centering
  \caption{Comparative study of the anomaly classification and segmentation performance on MVTec AD with various $\mu$ settings. Bold values indicate the best results.}
  \resizebox{\linewidth}{!}{%
    \begin{tabular}{@{}c|ccccccc@{}}
    \toprule[1.5pt]
    \multirow{2}{*}{$\mu$} & \multicolumn{4}{c}{Segmentation} & \multicolumn{3}{c}{Classification}\\
    \cmidrule(r){2-5}
    \cmidrule(r){6-8}
    ~ & AU-ROC& $F_1$-max& AP & AU-PRO& AU-ROC & $F_1$-max & AP \\
    \midrule
    0.47 & 93.1 & 48.0 & 44.7 & \textbf{88.3} & 93.3 & 94.2 & 97.1 \\
    0.50  & \textbf{93.2} & 48.1 & 44.8 & \textbf{88.3} & \textbf{93.4} & 94.3 & \textbf{97.2} \\
    0.53 & \textbf{93.2} & 48.1 & 44.9 & \textbf{88.3} & \textbf{93.4} & \textbf{94.4} & \textbf{97.2} \\
    0.55 & \textbf{93.2} & 48.1 & 44.9 & \textbf{88.3} & 93.3 & 94.3 & \textbf{97.2} \\
    \cellcolor{blue!10}0.57 & \cellcolor{blue!10}\textbf{93.2} & \cellcolor{blue!10}\textbf{48.2} & \cellcolor{blue!10}\textbf{45.0} & \cellcolor{blue!10}\textbf{88.3} & \cellcolor{blue!10}\textbf{93.4} & \cellcolor{blue!10}94.2 & \cellcolor{blue!10}\textbf{97.2} \\
    \bottomrule[1.5pt]
\end{tabular}
  }
  \label{table:mu}
\end{minipage}
\end{table*}

\subsection{Ablation Study}
Unless otherwise specified, all ablation experiments were conducted on the MVTec dataset.

\begin{figure}
    \centering
    \includegraphics[width=1\linewidth,trim=0 32 0 0, clip]{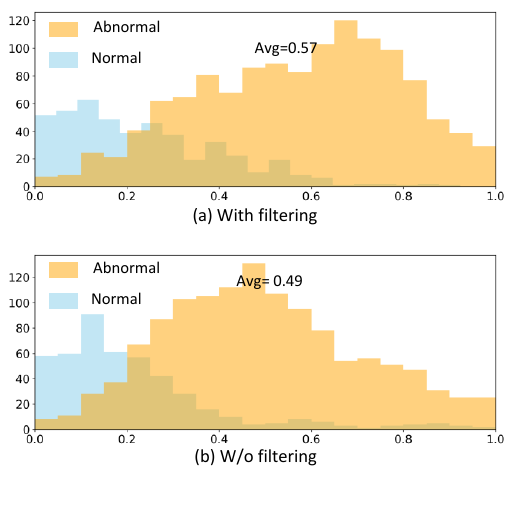}
    \caption{Histogram of anomaly classification scores on the MVTec AD dataset. (a) represents filtering out noisy features, (b) represents not filtering noisy features, and Avg denotes the average anomaly score.}
    \label{fig:filter}
\end{figure}

\noindent\textbf{Effectiveness comparison of noise filtering and different filtering methods.}
As shown in Fig.\ref{fig:filter}, we analyze the anomaly classification scores on the MVTec AD dataset\cite{bergmann2019mvtec} with and without noisy feature filtering, with 'Avg' denoting the average anomaly score before and after filtering. The figure shows a clear increase in anomaly scores following filtering, and the quantitative rise in the average score highlights the role of filtering in improving anomaly detection performance.

Additionally, we conduct experiments on different filtering methods. `W/o filtering' is used to indicate that no noise filtering is applied. As shown in Alg.\ref{alg:mutual-filtering}, this study implements a two-layer loop structure, where the outer loop iterates over neighborhood $F^r, r\in [1, 2,3]$, and the inner loop iterates over various layers $F^i, i \in [6, 12, 18, 24]$ for each $F^r$, denoted as $M^{r,i}$. The outer loop's neighbors, when set in reverse order, are denoted as $M^{r-,i}$. Switching the inner and outer loops is represented by $M^{i,r}$. The notation $M^{i*r}$ indicates a single-layer loop.

\noindent\textbf{Effectiveness comparison of different hyper-parameter.}
There are two hyperparameters in Eq.\ref{lamdda} and Eq.\ref{mu} that need to be tuned. This ablation experiment exclusively applies feature matching with noisy feature filtering from Sec.\ref{ficlip}, without incorporating anomaly score adjustments from image-text alignment. As shown in Tab.\ref{table:lambda} Tab.\ref{table:mu}, we conduct experiments with different values of $\lambda$ and $\mu$. The results indicate that the best performance is achieved when $\lambda$ is and $\mu$ is 2. Moreover, the parameters show only a minor effect on the experimental outcomes.
\begin{table}
  \centering
  \belowrulesep=0pt
  \aboverulesep=0pt
  \renewcommand{\arraystretch}{1.2}
  \caption{Experiments on different self-attention mechanisms. Bold values indicate the best results}
  \label{tab:attn}
  \resizebox{\linewidth}{!}{%
    \begin{tabular}{@{}c|ccccccc@{}}
    \toprule[1.5pt]
    \multirow{2}{*}{Attn} & \multicolumn{4}{c}{Segmentation} & \multicolumn{3}{c}{Classification}\\
    \cmidrule(r){2-5}
    \cmidrule(r){6-8}
    ~ & AU-ROC& $F_1$-max& AP & AU-PRO& AU-ROC & $F_1$-max & AP \\
    \midrule
    3        & 83.2          & 28.4          & 22.8          & 57.8          & 88.6          & 90.8          & 94.8          \\
    6        & 77.3          & 22.2          & 15.5          & 54.8          & 89.0          & 91.7          & 94.6          \\
    9        & \textbf{88.8} & 34.2          & 27.2          & 76.5          & 90.8          & 91.7          & 95.8          \\
    15       & 87.3          & 36.8          & \textbf{32.4} & 77.4          & 90.5          & 91.1          & 95.8          \\
    18       & 79.1          & 25.0          & 18.0          & 64.0          & 88.2          & 90.5          & 94.7          \\
    21       & 72.1          & 20.5          & 12.6          & 50.9          & 87.3          & 91.3          & 94.6          \\
    24       & 79.5          & 25.1          & 18.2          & 57.5          & 87.3          & 91.2          & 94.4          \\
    \midrule
    $v-v$      & 84.2          & 34.3          & 26.0          & 77.0          & 89.9          & 91.2          & 95.2          \\
    $q-q$      & 80.8          & 29.7          & 22.0          & 65.6          & 87.1          & 91.1          & 94.2          \\
    $k-k$      & 84.2          & 34.0          & 25.8          & 72.6          & 89.5          & 91.1          & 95.4          \\
    \midrule
    \cellcolor{blue!10}12 (ours) & \cellcolor{blue!10}88.6          & \cellcolor{blue!10}\textbf{37.5} & \cellcolor{blue!10}31.4  & \cellcolor{blue!10}\textbf{83.0} & \cellcolor{blue!10}\textbf{91.6} & \cellcolor{blue!10}\textbf{92.6} & \cellcolor{blue!10}\textbf{96.0}\\
    \bottomrule[1.5pt]
\end{tabular}
  }
  \vspace{-0.5em}
\end{table}

\noindent\textbf{Effectiveness comparison of different attention mechanisms.}
We achieve anomaly classification and segmentation by calculating the similarity between visual and text features. In the Tab.\ref{tab:attn}, different numbers denote the use of attention weights from the specified layer in place of the final layer’s weights, with $v-v$, $k-k$ and $q-q$ signifying modifications of the final layer’s $q-k$ attention to $v-v$, $k-k$, and $q-q$ attention, respectively. According to Tab.\ref{tab:attn}, replacing the final layer with the attention weights from the twelfth layer achieves the best results on most metrics. Additionally, as depicted in the Fig.\ref{fig:attn}, intermediate layer attention maps emphasize local semantics, whereas deeper layers focus on global tokens. Shallow layers, however, display a more chaotic structure and given that CLIP classifies using the final layer’s output, shallow layers may experience modality shifts.

\section{Conclusion} \label{sec:conclusion}
In this paper, we propose a new framework for zero-shot anomaly detection, named FiSeCLIP. We introduce an approach more aligned with practical industrial settings, using images within a batch as mutual references. Since directly using unlabeled test images as references can introduce noise, we fully leverage CLIP's multi-modal capabilities to filter out noisy features. We then obtain the anomaly score through feature matching. Additionally, we aggregate features with varying degrees of neighborhood overlap in a layered calculation, using the anomaly score at each degree to refine the precision of the previous filtering. Furthermore, we enhance CLIP’s self-attention mechanism by substituting the final layer’s attention weights with those of an intermediate layer that better retains local semantics. Comprehensive testing on anomaly detection datasets confirms that our model achieves SoTA performance.

\newpage
{
    \small
    \bibliographystyle{ieeenat_fullname}
    \bibliography{main}
}

\end{document}